\documentclass[12pt,psamsfonts,reqno]{amsart} 
\usepackage[foot]{amsaddr}

\usepackage{mathabx}
\usepackage{bbm}

\usepackage[margin=1in]{geometry}

\usepackage[svgnames]{xcolor}
\usepackage[pagebackref=true]{hyperref}
\hypersetup{colorlinks = true,extension = notused,linkcolor = DarkBlue,anchorcolor = black,citecolor = Navy,filecolor = black,pagecolor = black,urlcolor = blue}

\usepackage{todonotes}

\usepackage{tikz-cd}
\usepackage{epigraph}

\usepackage[numbers,sort&compress]{natbib}
\usepackage{hypernat}

\numberwithin{equation}{section}

\newcommand{\libm}[1]{\href{file://localhost/Users/pestun/Dropbox/lib/math/#1}{\nolinkurl{#1}}}

\newcommand{\BZ}{\mathbb{Z}}
\newcommand{\BC}{\mathbb{C}}
\newcommand{\BR}{\mathbb{R}}
\newcommand{\wo}{W}
\newcommand{\BW}{\mathbb{W}}
\newcommand{\BV}{\mathbb{V}}
\newcommand{\BP}{\mathbb{P}}
\newcommand{\sv}{\mathsf{v}}
\newcommand{\sw}{\mathsf{w}}
\newcommand{\si}{\mathsf{i}}
\renewcommand{\Re}{\operatorname{Re}}

\newcommand{\rank}{\operatorname{rank}}
\newcommand{\id}{\mathbbm{1}}
\newcommand{\Ve}{\mathsf{Vert}}
\newcommand{\Ed}{\mathsf{Edge}}
\newcommand{\In}{\mathsf{In}}
\newcommand{\Out}{\mathsf{Out}}

\usetikzlibrary{decorations.pathmorphing}

\begin{document}

\title {Tensor network language model}

\author{Vasily Pestun}
\author{Yiannis Vlassopoulos}

\address{Institut des Hautes \'Etudes Scientifiques (IH\'ES), Bures-sur-Yvette, France}
\email{  \{pestun,yvlassop\}@ihes.fr }

\begin{abstract}
  We propose a new statistical model suitable for machine learning of systems with long distance correlations such as natural languages.  
The model is based on directed acylic graph decorated
  by multi-linear tensor maps in the vertices and vector spaces in the edges,
  called tensor network. Such tensor networks have been previously employed 
  for effective numerical computation of the renormalization group flow on the
  space of effective quantum field theories and lattice models of statistical mechanics. 
  We provide explicit algebro-geometric analysis of the parameter moduli space for tree
  graphs,  discuss model properties and applications such as statistical translation. 
  \end{abstract}

\maketitle

\tableofcontents

\section{Introduction}
\epigraph{It must be recognized that the notion ”probability of a sentence” is an entirely useless one, under any known interpretation of this term.}{Noam Chomsky, 1969}

In natural language processing, unsupervised statistical learning of a language aims to
construct an efficient approximation to the probability measure 
on the set of expressions in the language learned from a sampling data set.

Currently, neural network models have proved to be the most efficient.
A particular success is attributed to models which construct
distributed word representations  \cite{Hinton_1986,Rumelhart_1988,Elman_1991},
that is a function $v: \wo \to \mathbb{V}$ from the set of words $\wo$ in
a language to a real vector space $\mathbb{V}$ of a certain dimension
depending on a particular model. 

Impressive results for constructing such a function $v$, also called continuous vector representation,
have been achieved in \cite{Mikolov_2013, Mikolov_2013a} (see also \cite{Bengio_2003} for
earlier construction of neural network language model and \cite{Schwenk_2007} for continuous space
word representations). A model in \cite{Mikolov_2013a} constructs
vector representation $v(w)$ of a word $w \in \wo$ by training a predictor
of words within a certain range of $w$ in the training sample of language. Curiously,
the function $v$ constructed in \cite{Mikolov_2013, Mikolov_2013a} was found to satisfy
interesting semantic and syntactic linear relations in English language such as 
\newcommand{\vs}[1]{v(\mathsf{#1})}
\[\vs{apples}-\vs{apple} \simeq \vs{cars}-\vs{car} \simeq  \vs{families}-\vs{family}\]
encoding the syntactic concept of grammatical number, 
\[\vs{Spain}-\vs{Madrid} \simeq \vs{France}-\vs{Paris} \simeq \vs{Italy} - \vs{Rome}\]
encoding the semantic concept of capital of a country,
\[\vs{king}-\vs{queen} \simeq \vs{man}-\vs{woman} \simeq \vs{uncle} - \vs{aunt}\]
encoding the semantic concept of gender.

The difficulties in the statistical modeling of language are due to the long 
range correlation. In \cite{Ebeling_1994} 
 mutual information\footnote{defined in terms of relative entropy} $I(l)$ between two characters in the text was measured 
as a function of the distance $l$ between positions of the characters
in two samples of English Literature (``Moby Dick'' by Melville and Grimm's Tales),
and it was found that $I(l)$ in the range $25 < l < 1000$ is well approximated by the power law 
\begin{equation}
\label{eq:Ipower}
  I(l) = c_1 l^{-\alpha} + c_2 
\end{equation}
with critical exponent $\alpha \simeq 0.37$. Further measurements in
\cite{Ebeling_1995} showed that the long distance correlation
are due to the structures beyond the sentence level.
In \cite{Altmann_2012} it was proposed to
explain the long range correlations in the text by a hierarchy of
levels in language which reminds the hierachical structure/renormalization group flow 
in the physical theories. The analysis in \cite{Montemurro_2002} confirmed long range correlations in
the sequence of integers constructed from the sequence of words in the text, 
where each word is mapped to a positive integer equal to the rank of this word
in the sorted list of individual word frequencies. Criticality
of language is not surprising with abundance of critical phenomena in
biology \cite{Mora_2011}.

Moreover, in \cite{Lin_2016} it was shown that a language
described by a formal probabilistic regular grammar
necessarily  has mutual entropy function $I(l)$ with exponentially fast decay
with respect to the distance $l$
\begin{equation}
\label{eq:Iexp}
  I(l) \simeq c \exp( - m l) 
\end{equation}
where $m$ is the inverse correlation length, or the mass gap in the physics terminology.
A formal probabilistic regular grammar is almost equivalent 
to a probabilistic finite automaton, hidden Markov model chain or
matrix product state called also tensor train decomposition in machine learning (see \cite{Vidal_2005a,Vidal_2005b} for more precise statements). These models can be thought as tensor networks
based on a graph with topology of 1-dimensional chain constructed from three-valent vertices
whose output edges correspond to the observables. 
In the context of language a matrix product state model is considered in \cite{MPS}. 
The absence of mass gap (power law correlation function, criticality, conformal structure, scale invariance)  observed in the natural language (\ref{eq:Ipower}) implies
that 1-dimensional tensor chain models (e.g. hidden Markov model or similar) 
fail to reproduce correctly the basic statistics of natural language: these models are  mass-gapped 
with exponential decay (\ref{eq:Iexp})
while a natural language is at criticality with the power law decay (\ref{eq:Ipower}).
This observation of \cite{Lin_2016} explains why hidden Markov models or similar 
 do not describe very well natural languages. 

On the other hand,  \cite{Lin_2016} also showed that 
a language described by a formal probabilistic context free grammar (probabilistic regular tree grammar)
has mutual entropy function function $I(l)$
of a power law type. A probabilistic regular tree grammar is modelled
by a tensor network based on a graph with topology of a tree in which leafs
correspond to an observable, see also \cite{Gallego_2017}. 
The hierarchy tree structure, for example in the case of a binary tree, means that from the state vectors of two words we construct the state vector of the \textit{sentence} of two words they form. From state vectors of sentences of two words we construct the state vector of the sentence of four words they form and so on. 

In physics this corresponds to the idea of iteratively coarse graining a system of many
locally interacting variables and the resulting
renormalization group flow on the space of theories developed by Kadanoff \cite{Kadanoff_1966},
Wilson \cite{Wilson_1971a,Wilson_1971b, Wilson_1975}, Fischer \cite{Fisher_1983}
and many others (see e.g. review \cite{Fischer_1998}). Further in
\cite{White_1992, White_1993} density matrix renormalization
algorithm was suggested which turned out to be quite efficient for numerical
solutions of quantum 1-D lattice systems. Algorithms implementing 
Kadanoff-Wilson-White real space renormalization
group on tensor tree networks have been developed in \cite{Shi_2005,Tagliacozzo_2009}.
See \cite{Silvi_2017} for a recent survey of tensor network models.

\begin{figure}
  \centering
  \includegraphics[width=5cm]{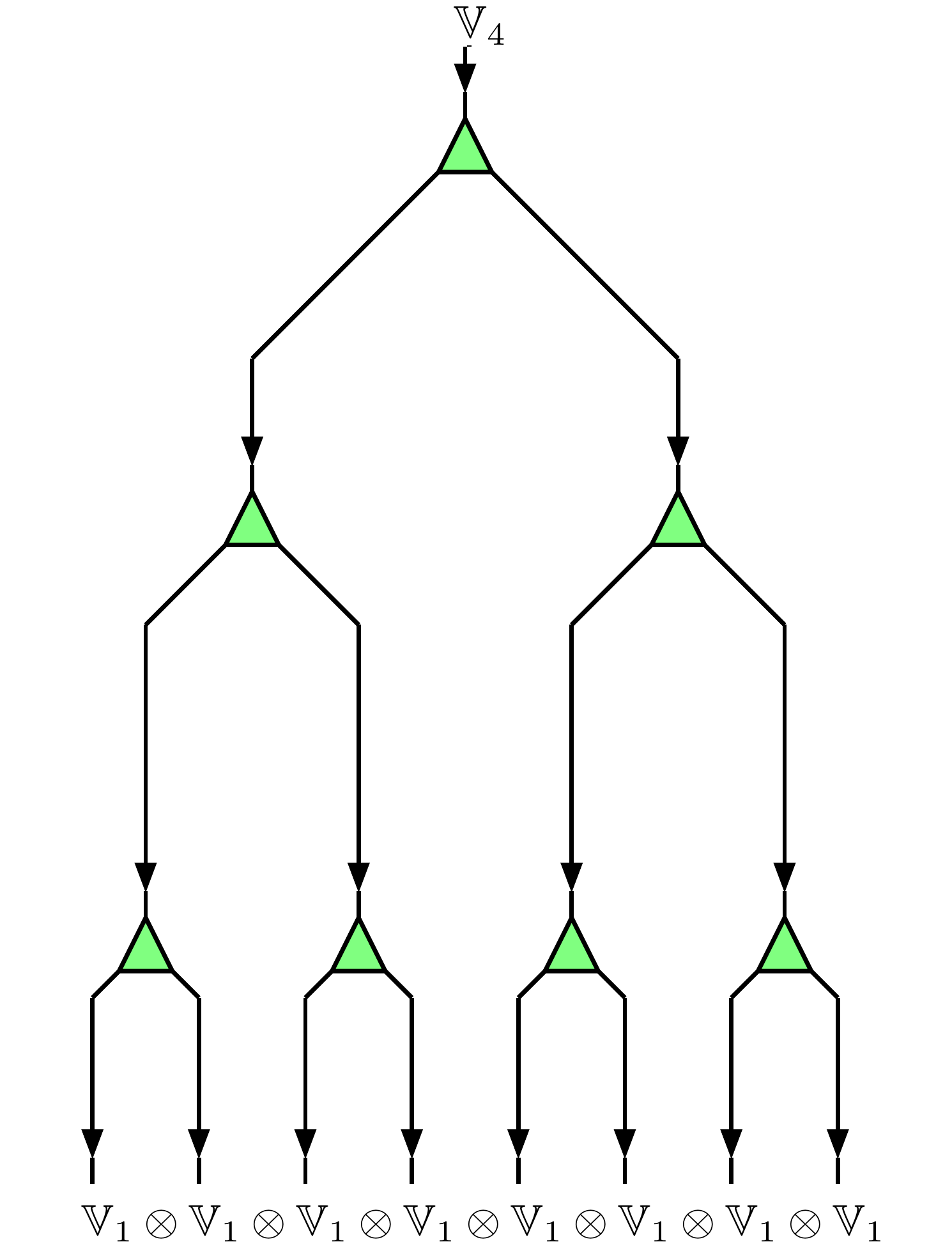} 
  \includegraphics[width=5cm]{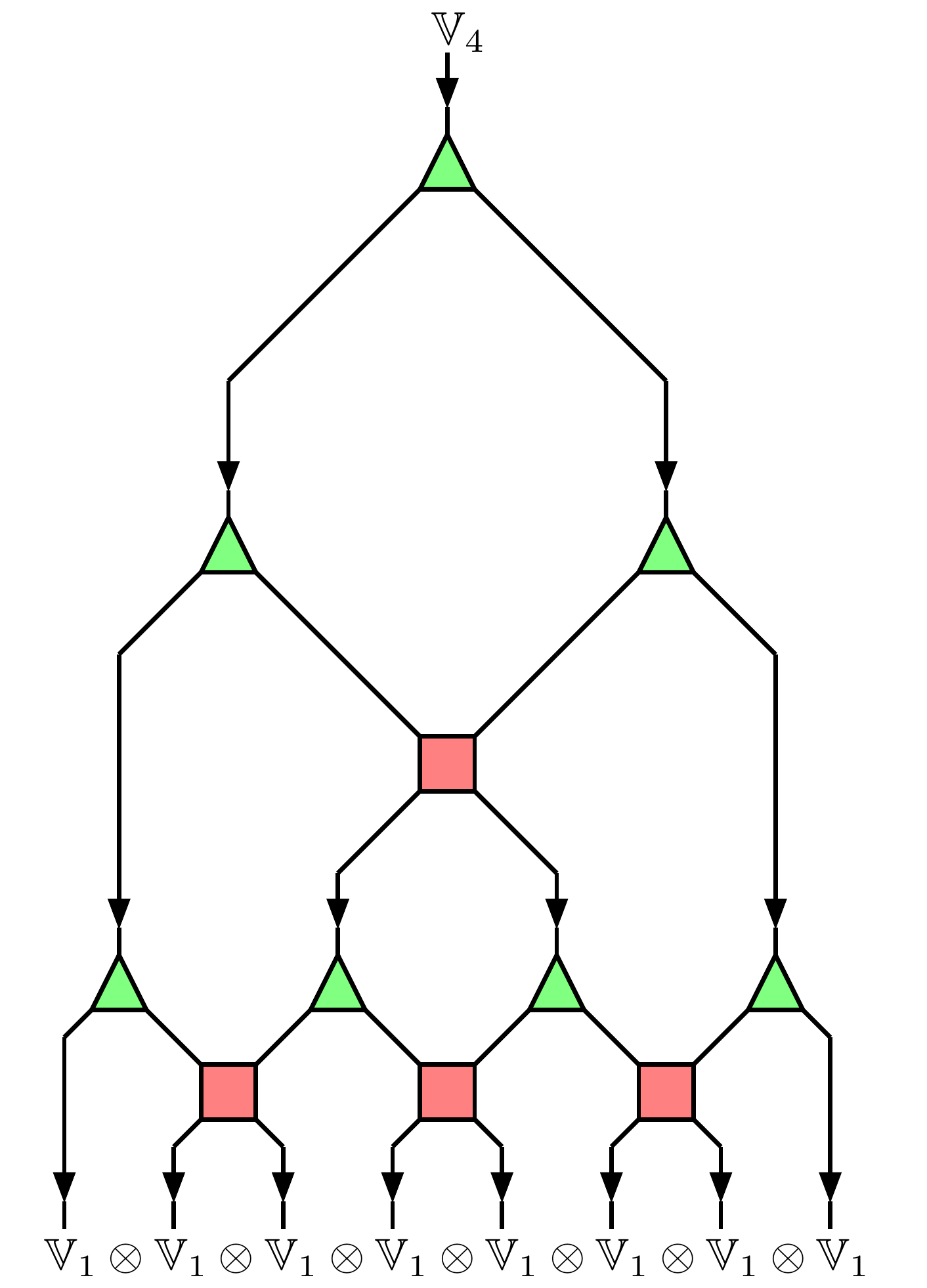}
  \caption{On the left: bare tree tensor network. On the right:
    MERA decorated tensor tree network is a map  $\BV_4 \to \BV_1^{\otimes 8}$
    obtained as composition of tensors at vertices, where $\BV_4$ is
    the vector space associated to the single incoming arrow at the top, and $\BV_1^{\otimes 8}$
    is the vector space associated to the 8 outgoing arrows at the bottom. Tensors of type
    (1,2) realizing maps  \[ \BV_{k+1}  \to \BV_{k} \otimes \BV_{k} \]
    are displayed by triangles, and disentagling tensors of type (2,2) realizing
    maps \[\BV_{k} \otimes \BV_{k} \to \BV_{k} \otimes \BV_k\] by squares. The direction
    of arrows is opposite to the renormalization group flow.
    The picture displays base layer of length $8$ and the binary tree. 
    In the conformal limit the length $n$ of base layer goes to infinity,
    and the number of layers scales as $\log n$. 
 \label{fig:mera8}}
\end{figure}

However, for critical systems the performance of bare tree tensor networks models
was limited due to effects of remaining long range entanglement. To handle
this entanglement the dimensions of Hilbert spaces in the layers
have to grow substantially when moving up
to the higher layers in which nodes represent compound objects.

In \cite{Vidal_2006} Vidal modified bare tensor tree network
by introducing disentangling operators between neighbor blocks
before applying each step of Kadanoff-Wilson-White  \cite{Wilson_1975,White_1993} density matrix renormalization group
projection operation, see Fig. \ref{fig:mera8}. Such tensor tree interlaced with disentaglers is called
Multiscale Entanglement Renormalization Ansatz, and it has been successfully
applied to study numerically many critical systems with impressive precision,
see e.g. review in \cite{Vidal_2009}. Intuitively, $p$-valent tree tensor networks 
could be thought as discrete models of AdS/CFT correspondence (holography) \cite{Maldacena:1997re,Gubser:1998bc,Witten:1998qj}, between the AdS side which is a discrete gravity theory
in discrete hyperbolic geometry represented by $p$-valent trees, and the CFT side
which is a critical (conformal) theory that lives on the boundary of the tree,
see e.g.\cite{Gubser:2016guj}. 


Motivated by
\begin{enumerate}
\item criticality of language \cite{Ebeling_1995,Ebeling_1994,Altmann_2012,Montemurro_2002,Mora_2011}
\item emergence of effective vector structure on the space
  of linguistic syntactic and semantic concepts \cite{Rumelhart_1988,Hinton_1986,Elman_1990,Mikolov_2013,Mikolov_2013a}
\item real space matrix density renormalization group or 
  discretized holographic AdS/CFT correspondence realized by tensor trees \cite{Kadanoff_1966,Wilson_1971a, White_1993,Shi_2005,Vidal_2006}
\end{enumerate}
in this paper we propose to use quantum MERA-like tensor networks for a statistical
model of language or other data sets with observed critical phenomena
and long range power law type correlation functions.

\subsection{Previous work}
Recurrent neural networks have been shown to have long range correlation \cite{Graves_2013,Lin_2016}. The connection between deep learning architectures and renormalization
group has been pointed out in several recent works \cite{Beny_2013,Mehta_2014, Lin_2016a}.
Moreover, in \cite{Cohen_2015,Levine_2017} it was shown
that an arithmetic circuit deep convolutional network is a particular case of a tree tensor network.  Linear matrix product states and density matrix renormalization group \cite{White_1992b} 
in the context of image recognition have been analyzed in \cite{Stoudenmire_2016}. On the other hand, deep neural networks in \cite{Torlai_2016} have been used
to compute correlation functions in the Ising model,
and \cite{Carleo_2016} deep neural networks
have been used to learn a wave-function of a quantum system. In \cite{Deng_2016} it was suggest to represent
topological states with long-range quantum entanglement by a neural network,
and in \cite{Huang_2017} it was suggested how to  accelerat Monte Carlo statistical
simulations with deep neural network.
In \cite{Chen_2017} some analysis has been put for  equivalence of
certain restricted Boltzmann machines and tensor network states.

While the present manuscript was in preparation we noticed works \cite{Han_2017,Liu_2017,Gallego_2017} which contain partial overlap with presented constructions.

\subsection{Acknowledgements} We would like to thank Maxim Kontsevich and John Terilla for useful discussions. The research of V.P. on this project has received funding from the European Research Council (ERC) under the European Union's Horizon 2020 research and innovation program (QUASIFT grant agreement 677368), and research of Y.V. received funding from Simons Foundation Award 385577.

\section{Quantum statistical models}
\subsection{Quantum states}

Let $\wo$ be the set of symbols from which a language is constructed.
For example, $\wo$ could be a set of words, a set of syllables, a set of ASCII characters,  a set $\{0,1\}$
in the binary representation, a set of musical characters, a set of tokens in computer programming
language, a set of DNA pairs  and so on. By $\sw = |\wo|$ we denote the number of symbols in the set $\wo$.

Let $\wo^{*}$ be the set of sequences of symbols in $\wo$, including the empty sequence
\begin{equation}
  \wo^{*} =\amalg_{n \in \BZ_{\geq 0}} \wo^{\times n} 
\end{equation}
where $\wo^{\times n} = \underbrace{\wo \times \wo \dots \times \wo}_{n}$.
An element of the set $S = \wo^{n}$ is a sequence of length $n$ consisting of symbols in $\wo$.

Let $\BW = \BC^{\wo} \simeq \BC^{\sw}$ be a vector space over the field of complex numbers
generated by $\wo$. 

Elements of $\BW$ are formal linear combinations of symbols in $\wo$ with complex coefficients. Using
Dirac bra-ket notations one can denote an element  $\psi \in \BW$ as
\begin{equation}
  \sum_{w \in \wo} \psi^{w} | w \rangle
\end{equation}
where $|w \rangle $ is a basis element in $\BW$ labelled by a symbol $w \in \wo$ and $\psi^{w}$
is a complex number.\footnote{We use the standard conventions to label coefficients of controvariant vectors $|\psi \rangle \in \BW$
by upper indices and coefficients of covariant vectors $ \langle \psi | \in \BW^{\vee}$ by lower indices}
   An element in
the vector space $\BW$ is called \emph{state}. 

For example, if the set of symbols $\wo$ is a set of English words, a state $\psi \in \BW$ could be equal to 
\begin{equation}
\psi = \frac{1  + \sqrt{3} \si }{4} | \mathsf{mountain} \rangle  + \frac{ \sqrt{3} }{2} |\mathsf{hill} \rangle
\end{equation}

We equip $\BW$ with Hermitian metric $\langle , \rangle$, that is a positive definite sesquilinear form, also called
inner product
\begin{equation}
\langle , \rangle:  \overline \BW \otimes \BW \to \BC
\end{equation}
where $\overline \BW$ denotes a vector space complex conjugate to $\BW$ in such a way
that $\wo$ is the standard basis 
\begin{equation}
  \langle w | w' \rangle = \delta^{w}_{w'}, \qquad w, w' \in \wo
\end{equation}
where 
\[\delta^{w}_{w'} =
\begin{cases}
  1, w = w' \\
  0, w \neq w'
\end{cases}\] is the Kronecker symbol. 
 In other words, $\BW$ is a finite-dimensional
 Hilbert space with orthonormal basis $\wo$.
 For every state $| \psi \rangle \in \BW$ there is an adjoint state $\langle \psi | \in \BW^{\vee}$ induced by the Hermitian
 metric $\langle , \rangle$.

For example, the norm of a state $|\psi \rangle = \sum_{w} \psi^{w} |w \rangle$ is
\begin{equation}
  \langle \psi |  \psi \rangle  = \sum_{w \in \wo} \bar \psi_{w} \psi^{w}
\end{equation}
where $\overline \psi_w$ denotes complex conjugation of a complex number $\psi_w$. 

To a length $n$ sequence of symbols $ s = (w_1,\dots, w_n) \in \wo^{\times n}$ we associate a basis element
$| s \rangle =  | w_1 w_2 \dots w_n \rangle$
in the vector space 
\begin{equation}
  \BW^{\otimes n} = \underbrace{\BW \otimes \BW \dots \otimes \BW}_{n}
\end{equation}

A generic state $\Psi \in \BW^{\otimes n}$ is a linear combination of basis elements with complex coefficients
\begin{equation}
  \Psi = \sum_{ s \in \wo^{n}} \Psi^{s} | s\rangle 
\end{equation}

An operator $o_s: \BW^{\otimes n} \to \BW^{\otimes n}$ defined as
\begin{equation}
  o_s =  |s \rangle \langle s|
\end{equation}
is the projection operator on the basis element $| s \rangle $. In particular,
\begin{equation}
  \langle \Psi o_s \Psi \rangle  =  \langle \Psi | s \rangle  \langle s | \Psi \rangle  = | \langle s | \Psi \rangle |^2 =
  \overline \Psi_s \Psi^{s}  
\end{equation}
is the absolute value square of the coefficient $\Psi_s$, thus it is a real non-negative number.
A state $\Psi$ is called normalized if it has unit norm
\begin{equation}
  \langle \Psi | \Psi \rangle   = 1
\end{equation}
For a normalized state $\Psi$ it holds that
\begin{equation}
 \sum_{s \in \wo^{\times n}} \langle \Psi o_s \Psi \rangle  = 1
\end{equation}

\subsection{Pure state statistical model}
A statistical model on the set $\wo^{\times n}$ is 
a family of probability distributions $\mu$ on $\wo^{\times n}$ fibered over 
the base space of parameters $\mathcal{U}$.
That is,  for each parameter  $u \in \mathcal{U}$ there is a positive real valued function 
\begin{equation}
\label{eq:statmodel}
  \mu_u: \wo^{\times n}  \to \BR_{\geq 0} , \qquad u \in \mathcal{U}
\end{equation}
such that
\begin{equation}
\label{eq:munorm}
  \sum_{s \in \wo^{\times n}} \mu_u(s) = 1, \qquad u \in \mathcal{U}
\end{equation}

The base space $\mathcal{U}$ of parameters can be thought as moduli space of distributions in a given statistical model.

A quantum (pure state) statistical model on $\wo^{\times n}$ with the space of parameters $\mathcal{U}$
is a family of normalized states $\Psi \in \BW^{\otimes n}$ fibered over the base space $\mathcal{U}$. 
That is for each point $u$ in the space of parameters $\mathcal{U}$ there is a normalized
state $\Psi_u \in \BW^{\otimes n}$:
\begin{equation}
\label{eq:Psinorm}
  \langle \Psi_u | \Psi_u \rangle = 1.
\end{equation}

A quantum statistical model $\Psi$ induces classical statistical model $\mu$ by the Born rule 
\begin{equation}
\label{eq:Born}  \mu(s) := \langle \Psi o_s \Psi \rangle 
\end{equation}
Indeed, $\mu(s)$ is a real non-negative number, and the normalization (\ref{eq:munorm}) follows
from (\ref{eq:Psinorm}).

We remark that in quantum physics with a Hilbert space of states $\mathcal{H}$,
a $\mathcal{U}$-family of normed  states $\Psi_u \in \mathcal{H}, \langle \Psi_u | \Psi_u \rangle  = 1 $ is often called  a  wave-function ansatz. A typical
problem posed in quantum physics is to find a ground state of a quantum system, that is 
an eigenstate with the lowest eigenvalue of a positive definite Hermitian operator $H$ (Hamiltonian)
acting on a Hilbert space $\mathcal{H}$. Often the exact solution of this problem is
not possible, and one resolves to approximate methods. A wave-function ansatz is such an approximate
method. While the exact ground state problem is equivalent to the minimization problem 
\begin{equation}
\min_{| \Psi \rangle \in \mathcal{H}, \langle \Psi \Psi \rangle = 1} \langle \Psi | H|  \Psi \rangle
\end{equation}
over the space of all states in $\mathcal{H}$ with unitary norm, an approximate solution by an ansatz 
searches the minimum over, usually,  much smaller subset of states $\{\Psi_u| u \in \mathcal{U}\}$
\begin{equation}
 \min_{u \in \mathcal{U}} \langle \Psi_u | H | \Psi_u \rangle
\end{equation}

Similarly, using the Born rule induction (\ref{eq:Born}) of a statistical model $\mu$ from a pure state model $\Psi$,
we can think about a family of quantum states $\{\Psi_u, u \in \mathcal{U}\}$ as a particular case of classical statistical model.

\subsection{Learning the model}
Given a statistical model $\mu_u$ on a discrete set $S$,  a statistical learning of a model is to find an optimal value $u_{*}$ of parameters in $\mathcal{U}$,
which means that $\mu_{u_*}$ approximates best an observed distribution $\mathring\mu$ in a sample of data points in $S$. 
  Formally, a sample of data 
  is a multiset 
  \[\mathcal{S} = (S, m: S \to \mathbb{Z}_{\geq 0})\]
 based on the set $S$, that is a set of pairs $(s, m(s))$ where $s \in S$ is an observed data point, and non-negative integer $m(s) \in \BZ_{ \geq 0}$ is the multiplicity
 of point $s$ in the sample.

 An observed distribution $\mathring{\mu}_{\mathcal{S}}$ associated to a sample $\mathcal{S}$ is a normalized frequency function $S \to \mathbb{R}_{\geq 0}$ defined by
 \begin{equation}
   \mathring{\mu}_{\mathcal{S}} (s) = \frac{m(s)}{|\mathcal{S}|}
 \end{equation}
 where $|\mathcal{S}|: = \sum_{s \in S} m(s) := \sum_{s \in \mathcal{S}} 1 $ is the cardinality of the sample $\mathcal{S}$. We use conventions where $\sum_{s \in \mathcal{S}}$ yields $s \in S$ with multiplicity
 $m(s)$.

 A distance between between two probability distributions $\mathring{\mu}$ and $\mu_{u}$
can be defined as Kullback-Leibler (KL) divergence 
 \begin{equation}
   D(\mathring{\mu}||\mu_u) = \sum_{s \in S} \mathring \mu(s) 
\log \frac{ \mathring{\mu}(s)}{\mu_u(s)} 
\end{equation}
This definition of distance between two probability
distributions satisfies certain natural axioms in the information theory that also
lead to the standard definition of the entropy of a distribution.

Hence, a standard method of learning a statistical model $\mu_u$ from a sample
of data points $\mathcal{S}$ is minimization of the KL divergence
between probablity distribution $\mu_{u}$ and the observed distribution $\mathring{\mu}$,
so that the optimal value of the parameters $u$ is 
\begin{equation}
  u_* = \arg \min_{u \in \mathcal{U}} D(\mathring{\mu}_\mathcal{S}||\mu_{u})
\end{equation}
Minimization of KL divergence bewteen $\mathring{\mu}$ and $\mu_u$ is equivalent
to maximization of log-likelihood $\sum_{s \in S} m(s) \log \mu_u(s)$,
that is
\begin{equation}
  u_{*} = \arg \max_{u \in \mathcal{U}} \sum_{s \in \mathcal{S}} \log \mu_u(s) 
\end{equation}
using the multiset yield notation $s \in \mathcal{S}$.

In particular, for a quantum pure state statistical model $\Psi_u$ and data sample $\mathcal{S}$, 
the optimal value of $u_{*} \in \mathcal{U}$ is
\begin{equation}
\label{eq:target}
 u_{*} = \arg \max_{u \in \mathcal{U}}   \sum_{s \in \mathcal{S}} (\log \Psi_u(s) + \log\overline \Psi_u(s))
\end{equation}

\section{Isometric tensor network model}

Now we consider a particular pure state statistical model $\Psi_u$ on
the Hilbert space $\mathcal{H} = \BW^{\otimes n}$ called isometric (or unitary) tensor
network model.

\subsection{Isometric maps}
Let $\BV, \BW$ be any Hermitian vector spaces, then a map
\begin{equation}
  u: \BV \to \BW
\end{equation}
is called isometric embedding (isometry for short) if the Hermitian metric on $\BV$ is equal to the pullback by $u$
of Hermtian metric on $\BW$. In other words, for any $v, v' \in \BV$ and $w = u v, w' = u v'$ 
it holds that
\begin{equation}
  \langle w  | w'  \rangle_{\BW} = \langle v | v'  \rangle_{\BV} 
\end{equation}
Since Hermitian metric is non-degenerate
\begin{equation}
  \rank u = \dim \BV 
\end{equation}
and isometric embedding exists only if $\dim  \BW \geq \dim \BV$.

An equivalent definition of isometric embedding of Hermitian spaces $u: \BV \to \BW$ is
that
\begin{equation}
\label{u-normed}
    u^{*} u = \id_{\BV}
\end{equation}
where $u^{*}: \BW \to \BV$ is the adjoint map and $\id_{\BV}$ is the identity map on $\BV$.
With respect to a basis on $\BV$ and a basis on $\BW$, the isometric property
of a map $u^{w}_{v}$ means 
\begin{equation}
  g_{\bar v v'} = \bar u^{\bar w}_{\bar v} u^{w'}_{v'} g_{\bar w w'}
\end{equation}
where $g_{\bar v v'}$ and $g_{\bar ww'}$ are components of Hermitian metric on $\BV$ and $\BW$.

The definition (\ref{u-normed}) also
implies that the operator $u u^{*}: \BW \to \BW$ is
a projection, since
\begin{equation}
( u  u^{*} )^2  = u u^{*} u u^{*} =  u u^{*} 
\end{equation}

We can think about morphism $u^{*}: \BW \to \BV$ as a projection on the image of $u$ in $\BW$.

A unitary transformation can be defined
as an isometry $\BV \to \BV$. The set of unitary transformations
of $\BV$ forms a group called unitary group $U(\sv)$. 
In this sense an isometry $\BV \to \BW$ is a
generalization of the notion of unitary transformation.

In general, the space $\mathcal{U}_{\BV, \BW}$ of isometries $\BV \to \BW$ is
not a group, but a homogeneous space isomorphic to the quotient
\begin{equation}
 \mathcal{U}_{\BV, \BW} =  \frac{U(\sw)}{U(\sw-\sv)}, \qquad \dim_{\BR} \mathcal{U}_{\BV, \BW} = 2 \sw \sv  - \sv^2
\end{equation}

For example, a normalized state $\psi \in \BW$ can be canonically identified
with an isometry $\tilde \psi: \BC \to \BW$ by taking the image
of $1 \in \BC$
\begin{equation}
  \psi  = \tilde \psi(1)
\end{equation}
and, indeed, the space of isometric embeddings of $\BC$ to $\BW$ is a unit
sphere $S^{2 \sw -1}$ isomorphic to the sphere of normalized states
\begin{equation}
  \mathcal{U}_{\BC, \BW} =  \frac{U(\sw)}{U(\sw-1)}  = S^{2 \sw  -1} 
\end{equation}
where
\begin{equation}
  S^{2 \sw -1} = \{ \psi \in \BW \quad | \quad   \langle \psi  \psi \rangle = 1 \}
\end{equation}

Also we define the reduced space $\tilde{\mathcal{U}}_{\BV, \BW}$ of
isometries from $\BV \to \BW$ that is obtained from $\mathcal{U}_{\BV, \BW}$ by
reduction by the unitary group $U(\sv)$ of automorphisms of the Hermitian space $\BV$.

For example, if $\BV = \BC$, then
\begin{equation}
  \tilde{\mathcal{U}}_{\BV, \BW} =  \frac{U(\sw)}{U(1) U(\sw-1)} = \BP^{\sw-1}
\end{equation}
which is a familiar statement from quantum physics
that the space of normalized states in~$\BW$ up to equivalence by phase rotation
forms the complex projective space $\BP^{\sw - 1}$. In generic case
\begin{equation}
\label{eq:GrassmanianU}
  \tilde{\mathcal{U}}_{\BV, \BW} =  \frac{U(\sw)}{U(\sv) U(\sw-\sv)} = \mathrm{Gr}_{\sv, \sw}, \qquad
  \dim_{\BC} \tilde{\mathcal{U}}_{\BV, \BW} = \sv \sw 
\end{equation}
where $\mathrm{Gr}_{\sw, \sw}$ denotes a Grassmanian of complex $\sv$-planes in $\sw$-dimensional
complex vector space.

The reduced space $\tilde{\mathcal{U}}_{\BV, \BW}$ is an example of moduli spaces of isometric tree tensor networks which
we discuss in more generality in section \ref{se:moduli}.

\subsection{Multi-linear isometric maps and tensor vertices}
For vector spaces $\BV_1, \BV_2, \dots, \BV_q$ and vector spaces $\BW_1, \BW_2, \dots, \BW_{p}$, a linear map $u$ of type $(p,q)$ is morphism
\begin{equation}
\label{eq:u-tensor}
  u: \BV^{\otimes q} \to  \BW^{\otimes p}
\end{equation}
where $\BV^{\otimes q}= \BV_1 \otimes \BV_2 \otimes \dots \otimes \BV_q$,
and  $\BW^{\otimes p} = \BW_1 \otimes \BW_2 \otimes \dots \otimes \BW_p$.

Let $W_i$ and $V_j$ be the bases of vector spaces $\BW_i$ and $\BV_j$ 
for $i \in [1,p]$ and $j \in [1,q]$. 
Then tensor $u^{w}_{v}$ of type $(p,q)$ is the table
of components  of $u$ represented as a $(p+q)$-dimensional
array table with $p$ upper indices
 and $q$ lower indices
\begin{equation}
 (u^{w}_{v}) =   ( u^{w_1 \dots w_p}_{v_1 \dots v_q}  \quad |  \qquad w \in W_1 \times \dots \times W_p, \qquad v \in V_1 \times \dots \times V_q )
\end{equation}

\begin{figure}
  \centering
  \includegraphics[width=5cm]{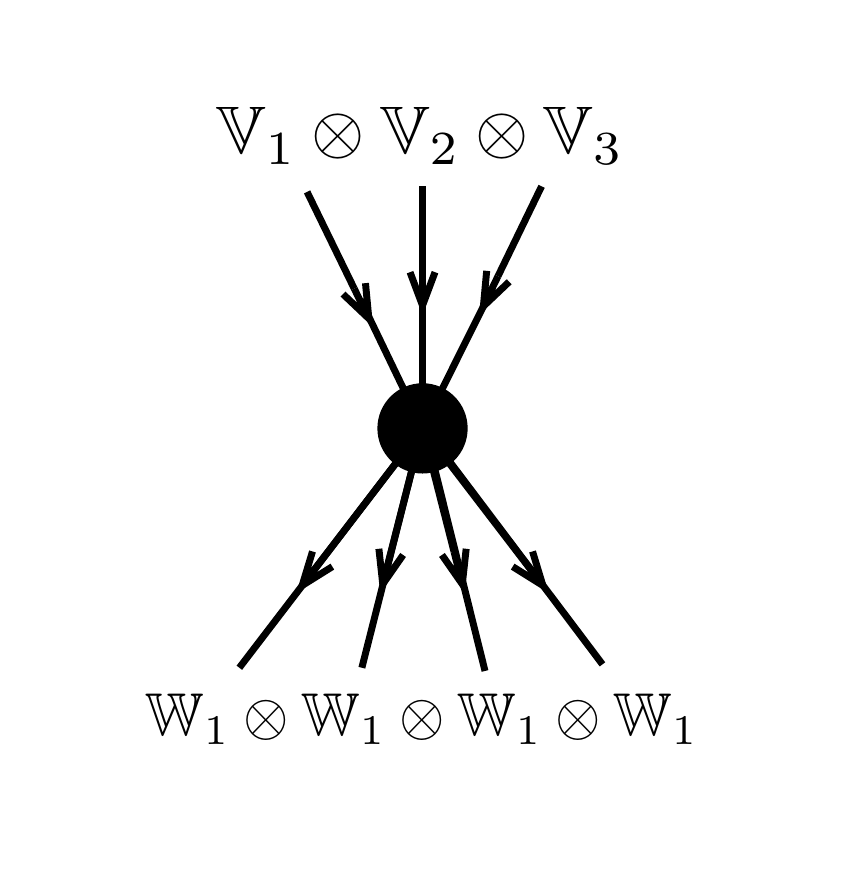}
  \caption{A tensor vertex $u: \BV^{\otimes 3} \to \BW^{\otimes 4}$ 
  }
  \label{fig:vertex}
\end{figure}

A Hermitian metric on $(\BW_i)_{i \in [1, p]}$ and on $(\BV_j)_{j \in [1, q]}$
induces Hermitian metric on $\BW^{\otimes p}$ and $\BV^{\otimes q}$, and
then a tensor $u^{w}_{v}$ is called an isometry if the map (\ref{eq:u-tensor})
is an isometry from $\BV^{\otimes q}$ to $\BW^{\otimes p}$ with respect
to the induced Hermitian metric. In the orthonormal basis this means
\begin{equation}
  \sum_{w \in W_1 \times \dots \times W_{p}} \bar u_{w}^{\tilde v} u^{w}_{v} =
  \delta^{\tilde v}_{v}, \qquad v, \tilde v \in V_1 \times \dots \times V_q 
\end{equation}
Graphically we display tensor vertex (\ref{eq:u-tensor}) as in figure \ref{fig:vertex}.

\subsection{Directed multigraphs}
Let $\gamma$ be a directed multigraph in which edges have
their identity (sometimes called quiver). More formally, a directed multigraph  $\gamma$ is $(\Ve, \Ed, s,t)$
where $\Ve$ is a set of vertices, $\Ed$ is a set of edges, $s: \Ed \to \Ve$ is a
source map which associates to each edge its source vertex,
and $t: \Ed \to \Ve$ is a target map which associates to each edge its target vertex.

A tensor network $U_{\gamma}$ (without inputs and outputs) is a decoration of a quiver $\gamma$
by a vector space $\BV_{e}$ at each edge $e \in \Ed$,
and linear map $u_{i}: \otimes_{e \in t^{-1}(i)} \BV_{e} \to \otimes_{e \in s^{-1}(i)} \BV_{e}$
at each vertex $i \in \Ve$.

Notice, that unlike the theory of quiver representations, in which
vertices are decorated by vector spaces, and the edges by maps between
vector spaces, in a tensor network edges of a quiver are decorated by vector spaces,
and the vertices are decorated by multi-linear maps from the
tensor product of vector spaces of incoming edges 
to the tensor product of vector space of outgoing edges. A vertex $i$ 
decorated by a linear map $u_i$ is called tensor vertex. 

To define an open tensor networ, or tensor network with a boundary
we add a set of external incoming edges and a set of external outgoing edges.
Formally, a quiver $\gamma$ with a boundary is $(\Ve, \Ed, \In, \Out, s, t)$
where $\Ve$ is a set of vertices, $\Ed$ is a set of internal edges,
$\In$ is a set of incoming edges, $\Out$ is a set of outgoing edges, 
$s: \Ed \sqcup \Out \to \Ve$ is a
source map which associates to each edge its source vertex,
and $t: \Ed \sqcup \In \to \Ve$ is a target map which associates to each edge its target vertex.

An open tensor network $U_\gamma = (\gamma, (u^{(i)})_{i \in \Ve}) $ is a quiver $\gamma$, possibly with a boundary,  in which each edge $e \in \Ed \sqcup \In \sqcup \Out$ is decorated
by a vector space $\BV_{e}$, 
and each vertex $i \in \Ve$ is decorated by a multi-linear map 
\begin{equation}
\label{eq:ui}
  u^{(i)}: \otimes_{e \in t^{-1}(i)} \BV_{e} \to \otimes_{e \in s^{-1}(i)} \BV_{e}
\end{equation}

An open tensor network $(\gamma, (u^{i})_{i \in \Ve})$  defines a
multi-linear map $u_\gamma$ called evaluation from the tensor product of vector spaces
in the $\In$ edges  to the tensor product of vector space in the $\Out$ edges
by composition of maps $u^{i}$ 
\begin{equation}
  u_\gamma : \bigotimes_{e \in \In} \BV_{e} \to \bigotimes_{e \in \Out} \BV_{e}
\end{equation}

If the boundary is empty, i.e. tensor network is closed, then $u_\gamma$ is a number. 

In components, evaluation map $u_\gamma$ is obtained by summing over all pairs of upper and lower indices
in the product of tensor vertices
\begin{equation}
\label{eq:ugamma}
  u_\gamma =  \sum_{ v \in \bigtimes_{e \in \Ed} V_{e}}  \, \,\prod_{i \in \Ve} (u^{i})^{v(s^{-1}(i))}_{v(t^{-1}(i))}
\end{equation}

We remark that a tensor network $(\gamma, U_\gamma)$ could be recognized
as a Feynman diagram of the directed graph $\gamma$ for a 0-dimensional
field theory with a pair of  fields $(\phi_e, \tilde \phi_e)$
with $\phi_e \in \BV_{e}$ and $\tilde \phi_{e} \in \BV^{\vee}_{e}$
for each edge $e \in \Ed$ with kinetic term $ \langle \tilde \phi_e, \phi_e \rangle$
and interaction tensor vertices (\ref{eq:ui})
\begin{equation}
  \langle  \prod_{e \in t^{-1}(i)} \tilde \phi,   u^{i} \prod_{e \in s^{-1}(i) } \phi_e \rangle
\end{equation}
Also, graphical representation of contraction of tensor indices
corresponding to the composition of multi-linear maps in vertices 
is known as Penrose graphical notation. 

\begin{figure}
  \centering
  \includegraphics[width=5cm]{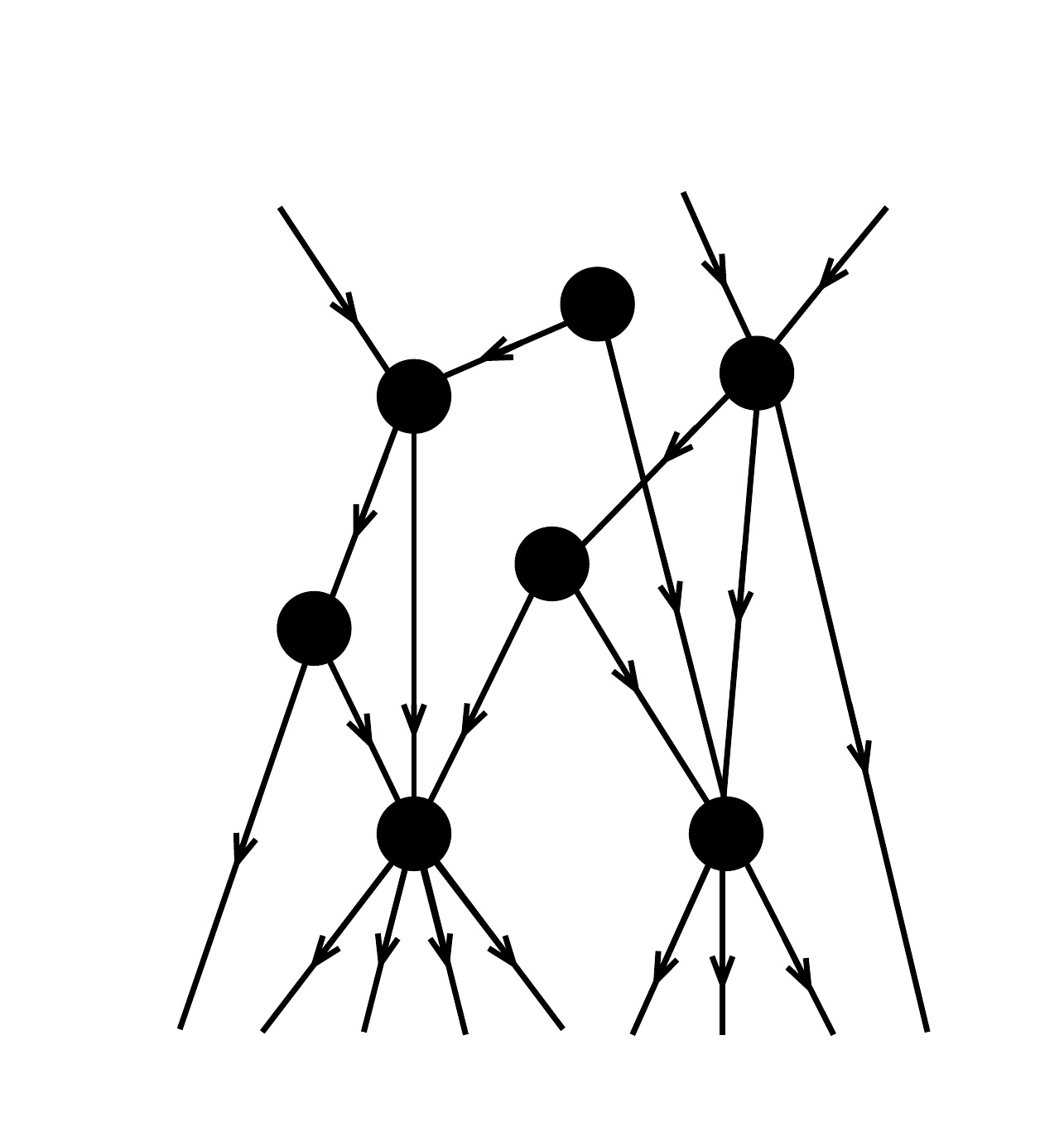}
  \caption{Directed acyclic graph}
  \label{fig:dag}
\end{figure}

\subsection{Directed acyclic graphs and props}

If we assume that directed multigraph $\gamma$ is acyclic (see Figure \ref{fig:dag}), i.e. $\gamma$
does not contain any directed cycles,
then mathematical structure that associates to $\gamma$ a
tensor network is called a colored prop \cite{Lawvere_1963,MacLane_1965,Joyal_1991,Yau_2008,Markl_2009,Hackney_2012,Yau_2015,Baez_2017,Yalin_2016}.   A prop is generalization
of the notion of operad. While operad takes several inputs and returns
a single output, a prop takes an
element of tensor product of several inputs and returns an element
in the tensor product of several outputs.
A tensor network for acyclic directed graph $\gamma$ is an object in
the endomorphism prop of a set of objects in
some symmetric monoidal category $\mathcal{C}$. 
In the above definition (\ref{eq:ui})~(\ref{eq:ugamma}) the category $\mathcal{C}$ 
pis a category of finite-dimensional vector spaces over complex numbers
with linear morphisms.

We remark that if directed multigraph $\gamma$ contains directed cycles, the corresponding
tensor network involves trace operation. Formally this
structure is encoded in the notion of colored wheeled prop.
In this situation $\mathcal{C}$ could
be any symmetric monoidal compact closed category so that
there is the trace operation in $\mathcal{C}$. Directed cycles in $\gamma$ correspond to the trace map and are called `wheels' in the context of
`wheeled prop'.
A category of finite-dimensional vector spaces with linear morphisms
has trace map, and thus it is suitable to build tensor network
(\ref{eq:ui})(\ref{eq:ugamma}) on arbitrary directed graph.

In the context of this paper we are interested in a
particular case of tensor networks called isometric tensor network.

An isometric tensor network $(\gamma, U_{\gamma})$
is a tensor network built on directed acyclic multigraph $\gamma$
in which edges are decorated by Hermitian vector spaces
and all multi-linear maps in vertices are isometric embeddings.
Hermitian vector spaces with isometric linear maps form a category,
since composition of isometries $f: \BV_1 \to \BV_2$ and $g: \BV_2 \to \BV_3$ is an isometry $  g \circ f: \BV_1 \to \BV_3$, and this category
is symmetric monoidal with the standard tensor product
of vector spaces. Abstractly, an isometric tensor network can be thought as an object in the endomorphism prop of a set of objects from the category of Hermitian
vector spaces with isometric morphisms.
Concretely, this means that if $\gamma$ is a directed acyclic multigraph
and all tensor vertices $u^{i}$ are isometries,
the evaluation map $u_\gamma$ (\ref{eq:ugamma}) is also an isometry.

\subsection{Isometric pure state tensor network model}

\begin{figure}
  \centering
  \includegraphics[width=5cm]{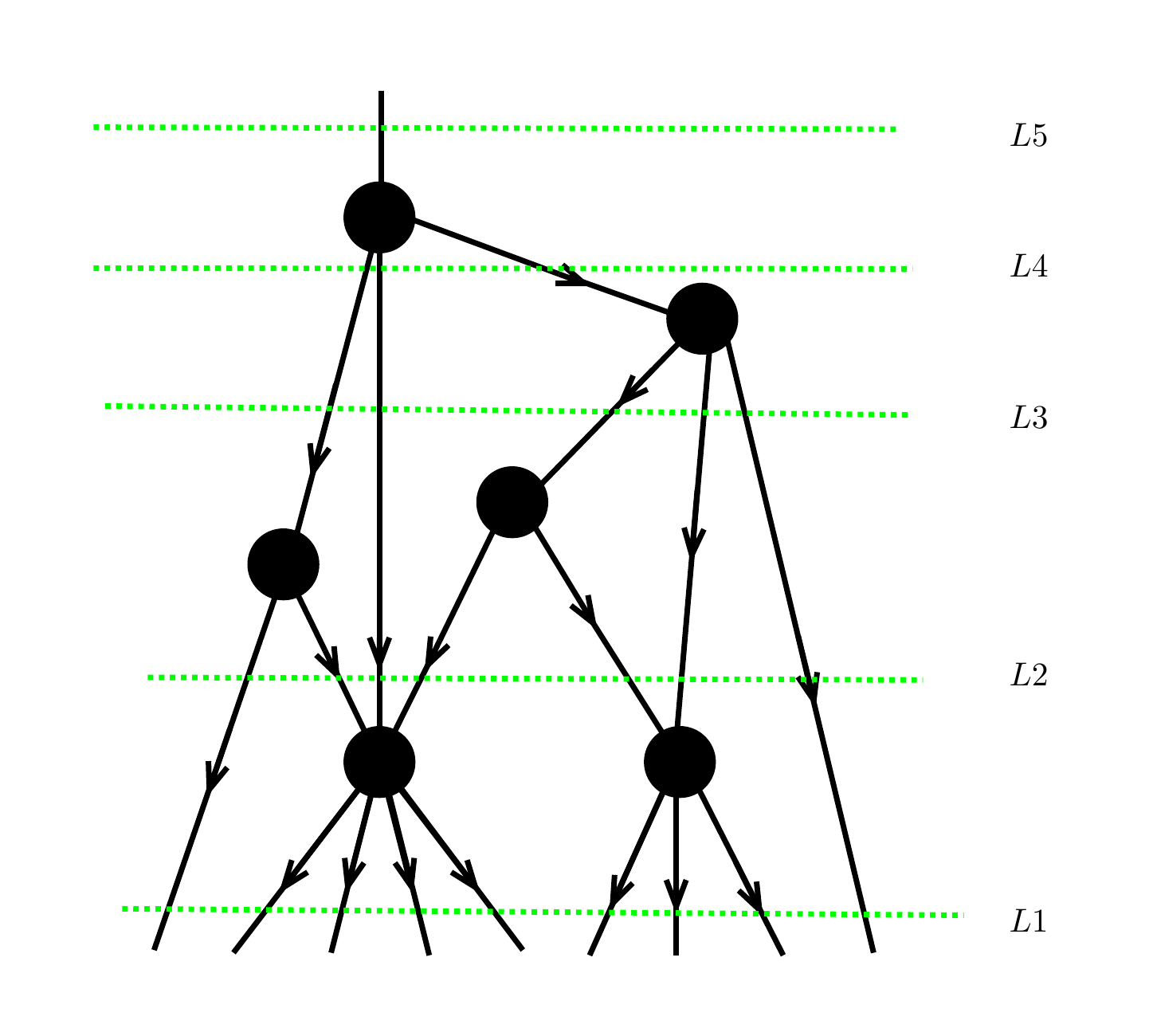}
  \caption{Pure state isometric tensor network model}
  \label{fig:network-single1}
\end{figure}

Given a Hilbert space $\mathcal{H}_n = \BW_1 \otimes \BW_2 \dots \otimes \BW_n $, whose
basis is a set $W^{n}$ of length $n$ sequences of symbols in $W$,  
a pure state isometric tensor network model $(\gamma, U_{\gamma})$ for a state 
\[
\Psi \in \BW_1 \otimes \BW_2 \dots \otimes \BW_n  
  \]
  is an isometric tensor network $(\gamma, U_\gamma): \BC \to \mathcal{H}$
  with a  single input vector space  $\BC$
  and output vector space $\mathcal{H}_n = \BW_1 \otimes \BW_2 \dots \otimes \BW_n$
  built on $n$ output edges, see
  
  The state $\Psi$ equals to the tensor network morphism $u_\gamma$ evaluated on $1 \in \BC$ 
  \begin{equation}
    \Psi = u_\gamma  1 
  \end{equation}

Notice, that in general, a directed acyclic graph $\gamma$ underlying a tensor network
is not a tree, i.e. there could be multiple directed paths from a node $i$ to a node $j$
(and `acyclic' here  means that directed cycles are not allowed).
In particular, MERA-like graph \cite{Evenbly_2011,Vidal_2009}
(see figure \ref{fig:mera8}) is directed acyclic but not a tree.

\begin{figure}
  \includegraphics[width=5cm]{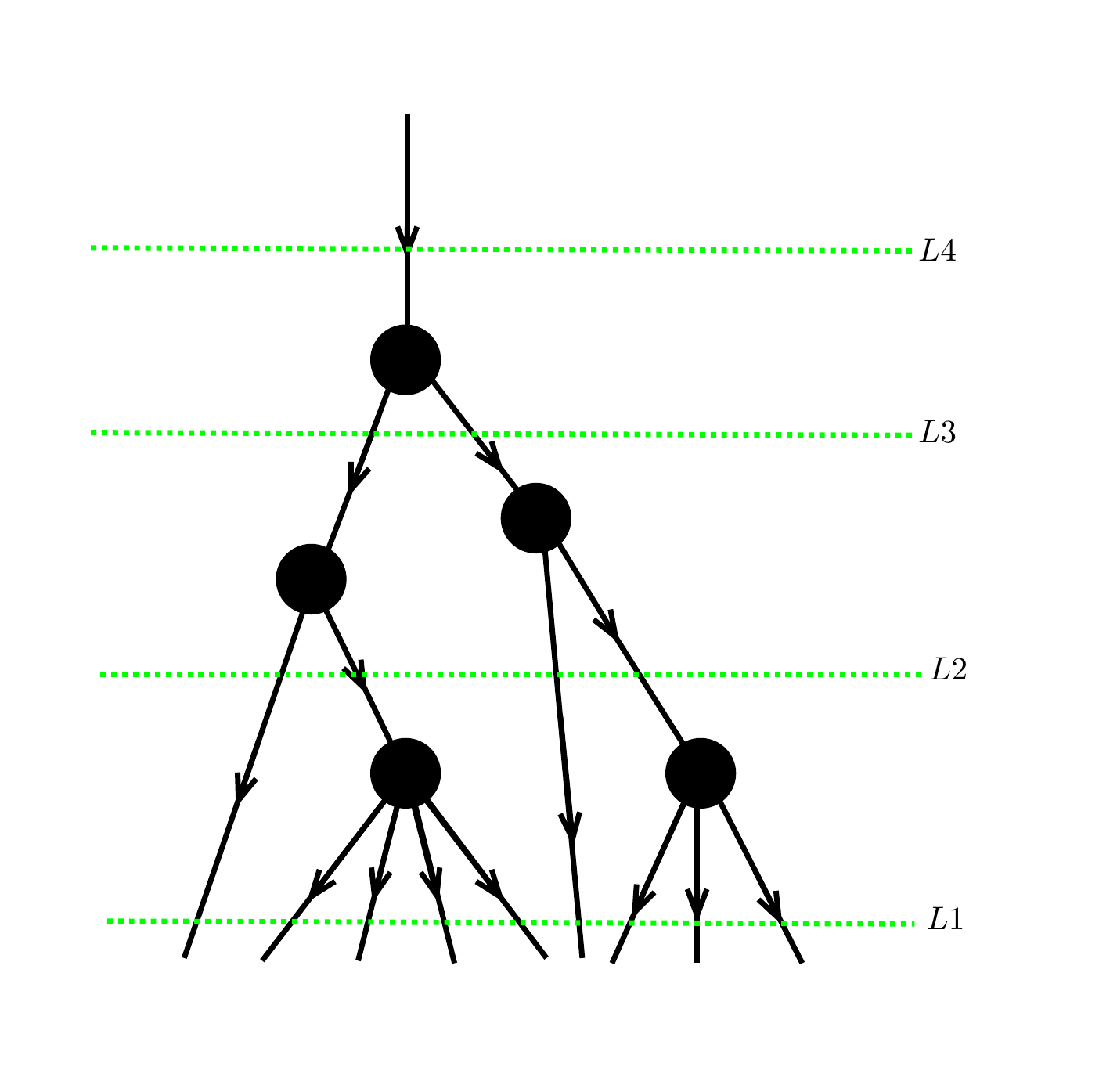}
  \caption{Pure state tree isometric tensor network model}
  \label{fig:tree-network}
\end{figure}

However, in a particular case
of isometric tree tensor network $(\gamma, U_\gamma)$, see Figure \ref{fig:tree-network}, 
it is computationally
easier to evaluate an amplitude of a sequence $s$
\begin{equation}
  \langle s | u_\gamma 1  \rangle  = \overline{\langle 1 u_\gamma^* s \rangle}
\end{equation}
since pulling back the state $\langle s | $ from the lower layer to the top vertex always
keeps it in the form of tensor product of states on the intermediate edges.
However, because of the observed criticality of language
the dimension of vector spaces at the top layers needs to grow for a sensible model.

A MERA-like graph with disentagling intermediate vertices $ V \otimes V \to V \otimes V$ is not a tree,
and therefore algorithms of the amplitude evaluation are computationally more costly. However,
it is feasible that the dimensions of the vector spaces in the edges at the higher layers of MERA graph
do not have to grow as fast \cite{Silvi_2017}, and MERA-like tensor model will turn out to be 
computationally more effective. 

\subsection{Slicing and layers} 
In any case,  a directed acyclic graph can be always sliced into layers $l$ (see
Figures \ref{fig:network-single1} and \ref{fig:tree-network}) such that
each layer contains vertices which are evaluated in parallel by tensor product. Denoting by
$[l]$ a set of vertices in layer $l$ we write
$u_{[l]} = \otimes_{i \in [l]} u^{(i)}$ for a morphism in the layer $l$, and then the evaluation
map of an isometric tensor network is a sequential composition of maps starting at the source (or higher) 
layer `L' with input $\BC$ and finishing at the target (or lower) layer $1$ with output $\mathcal{H}$. 
\begin{equation}
\label{eq:slicing}
  u_{\gamma,[1L]} =  u_{[12]} \circ \dots \circ u_{[L-1,L]}
\end{equation}

The adjoint map $u_\gamma^{*}: \mathcal{H} \to \BC$ is evaluated in the reverse order
\begin{equation}
    u_{\gamma,[L1]}^{*} =   u_{[L,L-1]}^{*} \dots u_{[21]}^{*} 
\end{equation}
where now the morphisms $u_{[l]}^{*}$ are not isometries in general, but projections if we identify the source
of $u_{[l]}$ with its image in the target space.

In particular, the projection operators $u_{[l]}^{*}$ do not preserve inner product. Consequently, by the analogy with renormalization group flow,  we expect that 
orthogonal commuting operators of projections on  basis states at the bottom layer (UV scale
in renormalization group/holography terminology),
such as projection operators $o_{[1]}$ 
\[ o_{[1]} = o_{| \mathsf{large} \rangle \otimes |\mathsf{hill}\rangle}, \qquad  o_{[1]}' = o_{| \mathsf{small} \rangle \otimes |\mathsf{mountain}\rangle} \]
are projected by the linear map to the similar operators $o_{l} \simeq o_{l}' $ operating at the middle layer $l$ 
\begin{equation}
o_{[l]} = u^{*}_{[lL]} o_{[1]} u_{[Ll]} , \qquad o_{[l]}' = u^{*}_{[lL]} o_{[1]}' u_{[Ll]}
\end{equation}

In the opposite direction, from intermediate higher layer $l$ to the bottom layer $1$, under the map 
\begin{equation}
n  o_{[1]}'' = u_{[1l]} o_{[l]} u_{[l1]}^{*} \simeq u_{[1l]} o_{[l]}' u_{[l1]}^{*}, 
\end{equation}
and in the context of the example we expect to see an operator of the form 
\begin{equation}
  o_{[1]}'' = o_{  c_1 | \mathsf{large} \rangle \otimes |\mathsf{hill}\rangle  +  c_2|  \mathsf{small} \rangle \otimes |\mathsf{mountain}\rangle}
\end{equation}
where $|c_1|^2, |c_2|^2$  are context-free probabilities of expressing similar concept with different words.

In other words, the context free choice between the UV layer expressions
$|\mathsf{large} \rangle \otimes |\mathsf{hill}\rangle $ vs $ | \mathsf{small} \rangle \otimes |\mathsf{mountain}\rangle$ is irrelevant at a higher level which operates within a Hilbert space $\mathcal{H}_{[l]}$ of higher level concepts.

Notice that the renormalization group flow preserves expectation values of relevant operators, in other words, if we define a state $ |\psi \rangle_l $ at the intermediate
level $l$ as
\begin{equation}
  | \psi \rangle_l  = u_{lL} | 1 \rangle 
\end{equation}
where
\begin{equation}
\label{eq:slicing-inter}
  u_{\gamma,[lL]} =  u_{[l,l+1]} \circ \dots \circ u_{[L-1,L]}
\end{equation}
then
\begin{equation}
  \langle \psi o_l  \psi\rangle_l  =   \langle \psi o_1  \psi \rangle_1
\end{equation}
where $o_l$ is an image under  renormalization group flow of the operator $o_1$ operating at the base layer `1'.

Therefore, the expectation value of $ o_{[1]}'' = o_{  c_1 | \mathsf{large} \rangle \otimes |\mathsf{hill}\rangle  +  c_2|  \mathsf{small} \rangle \otimes |\mathsf{mountain}\rangle}$
is approximately equal to the expectation value of $o_{[l]}$ or $o_{[l]}'$.

\section{Geometry of the moduli space}
\label{se:moduli}

Let $(\gamma, U_\gamma)$ be an isometric tensor network  built on a directed acyclic graph $\gamma$.
We define automorphism group (gauge group)
\begin{equation}
  \label{eq:Aut}
  \mathrm{Aut}(\gamma, U_\gamma)  = \prod_{e \in \Ed \sqcup \In} U(\BV_{e})
\end{equation}
to be the group of unitary transformations which act on all incoming and internal edges preserving
their Hermitian metric, so that tensor vertex $u^{(i)}$ transform as
\begin{equation}
  u^{(i)} \mapsto    \left( \prod_{e \in s^{-1}(i)} u_{e} \right)  u^{(i)} \left (\prod_{e \in t^{-1}(i)} u_{e}^{-1} \right)
\end{equation}
under the action of automorphism group element $(u_{e})_{e \in \Ed \sqcup \In}, (\id_{e})_{e \in \Out}$.

The moduli space of an isometric tensor network is defined as a quotient
\begin{equation}
  \mathcal{U}_{\gamma} = \oplus_{i \in \Ve} \mathrm{Isom} (\otimes_{e \in t^{-1}(i)} \BV_{e} ,  \otimes_{e \in s^{-1}(i)} \BV_{e}) \quad /\,  \prod_{e \in \Ed \sqcup \In} U(\BV_{e})
\end{equation}
where $\mathrm{Isom}(V, W)$ denotes the space of isometric maps from Hermitian space $V$ to Hermitian space $W$.

\subsection{Tree flag variety}
If $\gamma$ is a directed tree graph, the moduli space $\mathcal{U}_\gamma$ has a particular simple algebro-geometric
description. First notice, that the constraint that a map $u: \BV \to \BW$ is an isometry between Hermitian spaces $\BV$ and $\BW$ 
\begin{equation}
\label{eq:isom}
  u^{*}  u = \id_{\BV} 
\end{equation}
is a symplectic moment map for $U(\BW)$ group action on the Kahler space of all maps $\mathrm{Hom}(\BV, \BW)$.
By geometric invariant theory, the quotient of the level subset of the constraint
by the action of compact group is isomorphic to the quotient of the (stable locus) of the whole
set by the complexified group
\begin{equation}
\mathcal{U}_{u: \BV \to \BW} =   \{ u  \in \mathrm{Hom}(\BV,\BW) \quad | \quad u^{*} u = \id_{w} \} / U(\BV)
 \simeq    \mathrm{Hom}(\BV,\BW)^{\mathrm{stab}}  / GL(\BV)
\end{equation}
In the present case the stable locus $\mathrm{Hom}(\BV,\BW)^{\mathrm{stab}}$ is the space of injective maps,
and we obtain
\begin{equation}
  \mathcal{U}_{u: \BV \to \BW} \simeq \mathrm{Gr}_{\sv}(\BW)
\end{equation}
where $\mathrm{Gr}_{\sv}(\BW)$ denotes Grassmanian of $\sv$-dimensional planes in the vector space $\BW$,
with 
 $\dim_{\BC} \mathrm{Gr}_{\sv}(\BW) = \sv \sw - \sv^2 = (\sv - \sw) \sv$, c.f. (\ref{eq:GrassmanianU}). 

Now, if $\gamma$ is a directed tree with a single input, then each vertex $u^{(i)}$
has a single incoming input edge. Therefore, for a tree tensor network, the set of isometric constraints
on the maps $u^{(i)}$ in each  vertex $i$ (\ref{eq:isom}) 
is  a level set of symplectic moment map of the action 
of the full automorphism group (\ref{eq:Aut}). Consequently,
\begin{equation}
 \mathcal{U}_{\gamma_{\text{tree}}}   = \oplus_{i \in \BV} \mathrm{Hom}( \BV_{t^{-1}(i)}, \otimes_{e \in s^{-1}(i)} \BV_{e})^{\mathrm{stab}}/\prod_{i \in \Ve} GL(\BV_{t^{-1}(i)})
\end{equation}

In the simplest example,  when $\gamma$ is chain quiver in which each vertex has a single input
and a single output the moduli space $\mathcal{U}_{\gamma}$ is explicitly generalized
flag variety, i.e. the  moduli space of flags 
\begin{equation}
\BV_{\text{in}} \simeq \BV_{L} \subset \BV_{L-1} \subset \dots \subset \BV_{2} \subset \BV_{1} \simeq \BV_{\text{out}}
\end{equation}
which can be thought as $\mathrm{Gr}(\sv_L, \sv_{L-1})$ fibered over $\mathrm{Gr}(\sv_{L-1}, \sv_{L-2})$ fibered
over .... fibered over $\mathrm{Gr} (\sv_{2} , \BV_{1})$.

If $\gamma$ is a generic directed tree, the moduli space $\mathcal{U}_\gamma$ can be thought
as a generalization of flag variety for linear quiver, and could be called tree flag variety.

\begin{figure}
  \centering
  \includegraphics[width=5cm]{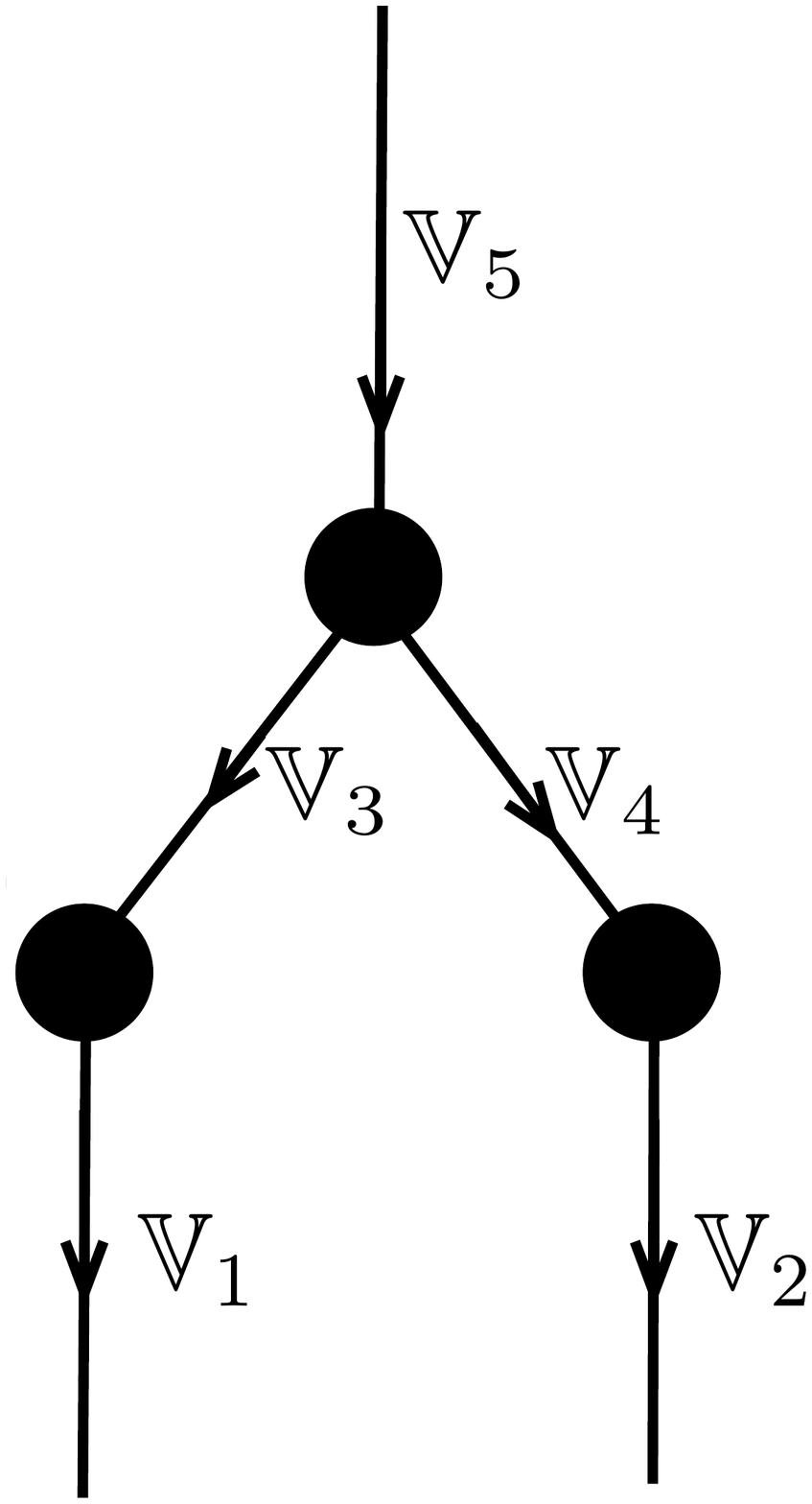}
  \caption{A small tree tensor network with moduli space $\mathcal{U}_\gamma$
    isomorphic to a tree flag variety given by a fibration
    with Grassmanian fibers 
    $\mathrm{Gr}(\sv_5, \mathcal{V}_3 \boxtimes \mathcal{V}_4)$
    over the base $\mathrm{Gr}(\sv_3,\BV_1) \times \mathrm{Gr}(\sv_4,\BV_2)$,
    where  $\mathcal{V}_3$, $\mathcal{V}_4$ are tautological vector bundles
    over the base. 
}
  \label{fig:treeflag}
\end{figure}

For the illustration, consider a tree isometric tensor network $\gamma$
displayed on Figure \ref{fig:treeflag}

In this case the moduli space $\mathcal{U}_\gamma$ is a fibration over the base 
$\mathrm{Gr}(\sv_3,\BV_1) \times \mathrm{Gr}(\sv_4,\BV_2)$ with the fibers $\mathrm{Gr}(\sv_5, \mathcal{V}_3 \boxtimes \mathcal{V}_4)$ where  $\mathcal{V}_3$, $\mathcal{V}_4$ denote tautological vector bundles over $\mathrm{Gr}(\sv_3,\BV_1)$ and $\mathrm{Gr}(\sv_4,\BV_2)$.

For general tree isometric tensor network $\gamma$, the moduli space $\mathcal{U}_\gamma$ is projective algebraic manifold
with has explicit description of a tower of fibrations, where 
the fiber at level $l$ is a product of Grassmanians, and then
the structure of fibration of products of Grassmanians at level $l+1$ uses external tensor product
of the tautological vector bundles of Grassmanians at level $l$ according to the combinatorics of tree vertices.

Hence, we deduce that for an isometric tree tensor network the parameter
moduli space $\mathcal{U}_\gamma$ is a smooth algebraic projective Kahler variety of complex dimension
\begin{equation}
  \dim_{\BC} \mathcal{U}_\gamma = \sum_{i \in \Ve}  \left( \sv_{t^{-1}(i)} \prod_{e \in s^{-1}(i)} \sv_{e}   - \sv_{t^{-1}(i)}^2\right)
\end{equation}

One can expect that the explicit algebro-geometric structure of the moduli space might be useful for
optimization algorithms that minimize the target function (\ref{eq:target}).

\section{Learning the network and sampling}
\subsection{Learning}
For an isometric tensor network $(\gamma, U_\gamma): \BC \to \mathcal{H}_n$,
where $\mathcal{H}_n = \BW^{\otimes n}$ is the Hilbert space based on length $n$ sequences, and a training
sample $\mathcal{S}$ of strings $s \in W^{\times n}$,  the effective free energy function $F(u)$ 
that needs to be minimized over the moduli space of parameters $u \in \mathcal{U}_\gamma$,
is equivalent to the KL divergence between
observed probability distribution and the model probability distribution (\ref{eq:target}) and is given by 
\begin{equation}
\label{eq:target1}
 F(u | \mathcal{S}) =  - \sum_{s \in \mathcal{S}} 2 \Re \log \langle s | u_\gamma | 1 \rangle 
\end{equation}
where $\langle, \rangle $ is the standard Hermitian metric on $\mathcal{H}_n$, and $| s \rangle $ is a basis
element in $\mathcal{H}_n$ labelled by a sequence $s \in W^{\times n}$. The summation over $s \in \mathcal{S}$
takes each element from the training multiset $\mathcal{S}$ with its multiplicity.

Since the objective function $F(u)$  is additive over the training sample $\mathcal{S}$,
a particular effective approximate algorithm to minimize $F(u)$ with a large sample $\mathcal{S}$ is a 
term-wise local gradient descent (called sometimes stochastic gradient descent). This algorithm constructs a flow on the moduli space $\mathcal{U}_\gamma$, where  each step
of the flow for a limited time (called learning rate) follows the gradient flow
associated to a single term $s$ in the objective function, so that evolution of $u$ for a step $s$ is
\begin{equation}
\label{eq:gradient1}
  \partial_{t} u  = - \nabla F (u | s), \qquad t \in [0, \eta]
\end{equation}
where $F(u|s)$ is 
\begin{equation}
  F(u|s) =  - 2 \Re \log \langle s | u_\gamma | 1 \rangle 
\end{equation}

Other learning methods based on the recursive
singular value decomposition of effecive density matrices developed in tensor networks
applied to many body quantum systems  \cite{Vidal_2009,Evenbly_2011,Vidal_2006,Lee_2015} might turn out also to be effective.

\subsection{Sampling}
To sample from a probability distribution of learned isometric tensor network a standard recursive
procedure can be used.

Namely, a sequence $s \in \mathcal{W}^{n}$ is sampled recursively according to the following algorithm.

The $k$-th element of $s$ is recursively sampled from the probability distribution of the $k$-th element
conditioned on the previously sampled $k-1$ elements of $s$, that is
\begin{equation}
  \mathrm{prob}(s_k | s_1 \dots s_{k-1})  = \frac{\langle \Psi o_{s_1 \dots s_k} \Psi \rangle }{\langle \Psi o_{s_1 \dots s_{k-1}} \Psi \rangle }
\end{equation}
where $o_{s_1 \dots s_k}$ denotes the projection operator on a length $n$ sequence with first $k$ elements
fixed to be $s_1 \dots s_k$,
i.e.
\begin{equation}
  o_{s_1 \dots s_k}  = o_{s_1} \otimes \dots \otimes o_{s_k}\otimes \id_{k+1} \otimes  \dots \otimes \id_{n}
\end{equation}

See also \cite{Ferris_2012} addressing sampling in the context of tensor networks. 

For tree tensor networks, the state $\psi_s = \langle s | u_\gamma |1 \rangle$ can be evaluated particularly
effectively by recursive composition over a leaf of the tree and deleting that leaf. The local
term gradient (\ref{eq:target1}) is also efficiently evaluated because of
the product structure of the evaluation morphism (\ref{eq:ugamma}). 
Namely, the gradient components for the  moduli of parameters in vertex $i$ is computed as 
pulling the tangent bundle for local variation $u^{(i)}$ in (\ref{eq:ugamma}) along the composition map
in (\ref{eq:ugamma}) of all remaining vertices (the pullback or `chain rule' for differential
of composition of functions is sometimes referred as `back propagation' in the context of neural network
optimizations).

\section{Discussion}

\subsection{Supervised model  and classification tasks}
A `supervised' version of algorithm is also possible, where a `supervised' label is
a relevant operator which survives at the higher (IR) levels of the network, such
as the general topic of the input text or other general feature relevant operators.
We simply add another leaf input to the network at higher level
decorated by a vector space whose basis is the set of higher level labels.

\subsection{Translation of natural languages}

It is expected that various human natural languages are in the same critical universality class $\mathcal{L}$
at the sufficiently high level of the network (i.e. at the sufficiently IR scale of
the renormalization group flow).

Then a translation engine from language $L_1$ to language $L_2$ can be constructed by connecting by unitary transformation  $S_{21}$ at
 sufficiently high layers $\mathcal{L}_1$ and $\mathcal{L}_2$ of two isometric tensor networks describing language $L_1$ and language $L_2$

\begin{equation}
\begin{tikzcd}
  \arrow{d}{u_2} \mathcal{L}_2  \arrow[leftsquigarrow]{r}{S_{21}}&   \mathcal{L}_1 \arrow{d}{u_1}\\
  L_2 &      L_1 
\end{tikzcd}
\label{eq:translation-map}
\end{equation}

A sequence $| s \rangle_1$ in language $L_1$ is translated
to a state $|\psi\rangle_2$ in the Hilbert space $\mathcal{H}_{L_2}$ of language $L_2$ equal to \begin{equation}
 |\psi\rangle_2   =  u_2 S_{21} u_1^{*} | s \rangle_1 
\end{equation}

The state $\psi \in \mathcal{H}_{L_2}$, in general, is a not basis element corresponding to a single  sequence
in $L_2$, but rather a linear combination
\begin{equation}
 | \psi\rangle_2 = \sum_{s \in W_2^{\times n}}  \psi_s | s \rangle_2 
\end{equation}
with complex coefficients $\psi_s$. If the state $|\psi\rangle_2$ is sampled, a basis sequence $|s\rangle_2$ from language $L_2$ is generated
with probability $| \psi_s |^2$. As expected, the translation (\ref{eq:translation-map}) is not isomorphism between languages in the base layer.
However, one can conjecture approximate isomorphism, or
a single universality class of all human languages at some deeper scale of renormalization group flow $\mathcal{L}_{1} \simeq \mathcal{L}_{2}$ that could be called the scale
of `meaning' or `thought'. The renormalization group flow $u_1^{*}$ from the base layer
$L_1$ to  the `meaning' layer $\mathcal{L}_1$ is many to one,
projecting different phrases in $L_1$  with equivalent meaning to the same state in
$\mathcal{H}_{\mathcal{L}_1}$. The map $u_2$ from the scale
of meaning $\mathcal{L}_2$ to the base layer of language $L_2$ is an 
ordinary isometry map between vector spaces, however,
in general, the expansion of the image state $|\psi\rangle_2$ 
over the basis in $L_2$ contains many phrases $|s \rangle_2$ weighted with
probability amplitudes. Each of these phrase is a possible translation
with corresponding probability $|\psi_s|^2$. 

\subsection{Network architecture}
In this note, for simplicity we have assumed a certain fixed topology of an isometric tensor network. However,
we expect that there is a natural generalization of the construction in which the topology of the underlying graph
of the model is not fixed, but arbitrary, so that the amplitudes $\psi_s$ are computed in the spirit of Feynman diagrams
where different graph topologies appear.

\subsection{Testing the model}
The presented construction is theoretical. It would be very interesting to implement the suggested models
and study its performance on various types of languages that display critical properties (human natural languages,
DNA sequences, musical scores, etc).

\bibliographystyle{utphys}
\bibliography{references}

\end{document}